%% file: ijcai22.tex
\documentclass{article}
\usepackage{ijcai22}

\usepackage{times}
\usepackage{soul}
\usepackage{url}
\usepackage[hidelinks]{hyperref}
\usepackage[utf8]{inputenc}
\usepackage[small]{caption}
\usepackage{graphicx}
\usepackage{amsmath}
\usepackage{amsthm}
\usepackage{booktabs}
\usepackage{algorithm}
\usepackage{algorithmic}
\usepackage{multirow}
\usepackage{tabularx}
\usepackage{amsfonts} 
\usepackage{makecell}
\newtheorem{definition}{Definition}

\title{Can We Find Neurons that Cause Unrealistic Images \\
in Deep Generative Networks?}

\author{
Hwanil Choi$^{1}$\and
Wonjoon Chang$^{1}$\And
Jaesik Choi$^{1,2,}$\footnote{Corresponding Author}\\
\affiliations
$^1$Graduate School of AI, KAIST, Seongnam, Republic of Korea\\
$^2$INEEJI, Republic of Korea\\
\emails
\{hwanil.choi, one\_jj, jaesik.choi\}@kaist.ac.kr
}

\begin{document}

\maketitle

\begin{abstract}

Even though Generative Adversarial Networks (GANs) have shown a remarkable ability to generate high-quality images, GANs do not always guarantee the generation of photorealistic images. Occasionally, they generate images that have defective or unnatural objects, which are referred to as `artifacts'. Research to investigate why these artifacts emerge and how they can be detected and removed has yet to be sufficiently carried out. To analyze this, we first hypothesize that rarely activated neurons and frequently activated neurons have different purposes and responsibilities for the progress of generating images. In this study, by analyzing the statistics and the roles for those neurons, we empirically show that rarely activated neurons are related to the failure results of making diverse objects and inducing artifacts. In addition, we suggest a correction method, called `Sequential Ablation’, to repair the defective part of the generated images without high computational cost and manual efforts.

\end{abstract}

\section{Introduction}

Generative Adversarial Networks (GANs) have been exhibited excellent performance for creating photo-realistic images, such as human faces, cars, and buildings.~\cite{karras2019style,brock2018large,karras2020analyzing,att2021,Esser_2021_CVPR,karras2021alias}. Additionally, GANs have been used to synthesize a new realistic image from several real images. For example, a target human image can be transformed to obtain another’s face~\cite{Kim_2021_CVPR}. These various uses of GANs have been utilized in many applications~\cite{DBLP:conf/ismir/HungCYY21,att2021,NEURIPS2021_f8417d04,wang2021dance}.

\begin{figure}[t!]
\centering
\includegraphics[width=1\columnwidth]{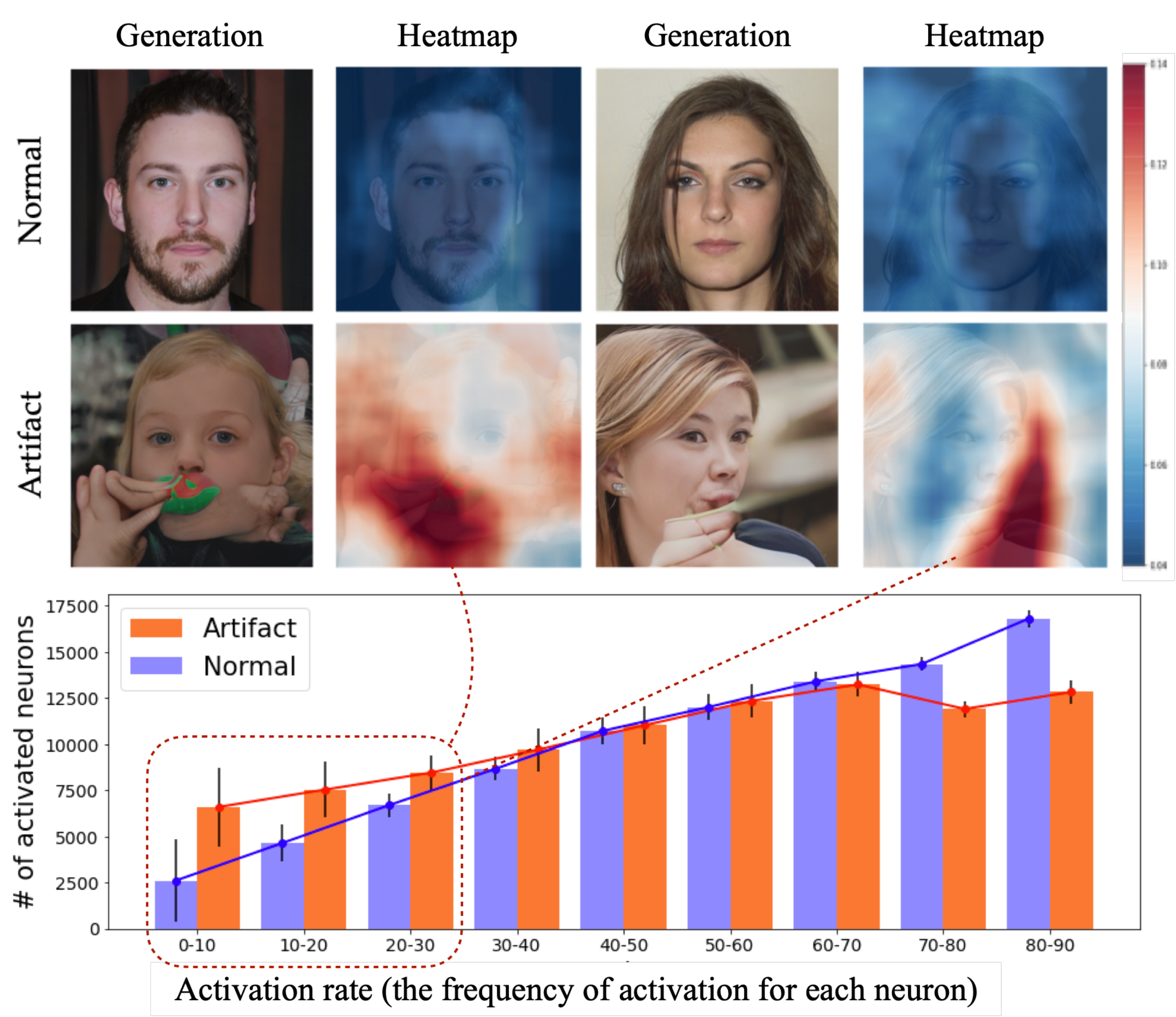} 
\caption{Heatmaps of rarely activated neurons for generated images. The first and second rows show the discrepancy between normal and artifact images respectively. The second and fourth columns describe the ratio of rarely activated neurons in featuremaps. In the red-colored area, where rarely activated neurons are activated a lot, there exist defective parts in the generated images. The bar plot shows that artifact images are highly related to rarely activated neurons compared to well-generated images.}
\label{fig:heatmap_norm_arti}
\end{figure}

Even though it is known that GANs have sufficient representative capacity to generate high-quality images, recent studies show that GANs still produce unrealistic images~\cite{odena2016deconvolution,bau2018visualizing} or only generate limited types of samples. In this context, even PGGAN~\cite{karras2018progressive} and StyleGAN~\cite{karras2020analyzing}, which are recognized as the high-performance models, may generate unrealistic images that include defective parts or unnatural objects, referred to as `artifacts'. As long as artifacts in generated images are not detected and fixed, the application of GANs will be limited.

Recently, there have been several attempts to identify the artifacts. GAN-Dissection~\cite{bau2018visualizing} analyzes internal units in a pre-trained GAN to identify which units contribute to generating a specific object in a generated image. In this work, the authors found several units that are responsible for typical artifacts. One method to identify the cause of an artifact is to find manually corresponding units creating artifacts. Another method for automatic identification of artifacts units is using unit-specific FID scores. They then turn off each identified unit, which means that activation values of all neurons in the selected unit are set to zero. Automatic Correction~\cite{tousi2021automatic} exploits an auxiliary classifier that is trained to distinguish artifacts from normal images and identifies the locations of defective parts by applying Grad-CAM~\cite{2019}. Handcrafted labels are required for fine-tuning this classifier. However, these studies lack analysis on the root cause of artifacts from the perspective of the generating process and they require manual sorting or additional training costs.

In this work, we focus on revealing which neurons make images distorted, how artifacts can be detected, and under which condition can make high quality natural images. In particular, we analyze PGGAN and StyleGAN2, which are known as high-performing models and have been widely adopted as the basic structures of generative models for high-quality high-resolution~\cite{10.1145/3470848,Kim_2021_CVPR,Yao_2021_ICCV}. We first analyze the frequency of activated neurons to reveal the following behavior of the generating process. Figure~\ref{fig:heatmap_norm_arti} illustrates the relation between defective parts of generated images and rarely activated neurons, which we will define as `low frequency rate'. We then empirically verify that defective parts in the artifact images are corresponding to neurons with low frequency rates. From this analysis, we show that artifacts can be censored without any additional training cost. Furthermore, we propose an ablation method to remove defective parts in an unsupervised manner. This method doesn't require an additional training procedure. This experimental results support our hypothesis both qualitatively and quantitatively. The main contributions of our work are as follows: 
\begin{itemize}
    \item We analyze the frequency of activation in feature maps in order to reveal the properties of the generating procedure.
    \item We experimentally show that not only are the artifacts related to the rarely activated neurons, but also are the artifacts detected by this property.
    \item We provide qualitative and quantitative results. Furthermore, our analysis can be applied to detect and remove artifacts from generated images.
\end{itemize}

\section{Related work}

\subsection{Generative Adversarial Networks}
GANs~\cite{goodfellow2014generative} have been leading a new paradigm of generative models. A GAN consists of a discriminator and a generator. The discriminator classifies between two images distributions: real images and fake images. The generator creates fake images to deceive the discriminator. One improved version of conventional GAN structure is Progressive Growing GAN~\cite{karras2018progressive}, which is trained sequentially by increasing the resolution, and has emerged as a high-quality models. While recent models based on the GAN structure produce indistinguishable photo-realistic images, there have been few studies on the inner process of the GAN model.

\subsection{Style-based GAN}
StyleGAN~\cite{karras2019style} was proposed as a novel generating mechanism for GAN to generate high quality and resolution images. This model has two parts: a mapping network and a style generator. The former transforms an input latent code $z$ to a style code $w$. The latter generates an image with the trained constant tensor and the style code $w$. Although StyleGAN affords good performance, blob artifacts can sometimes be found in the images. To address this, they suggested an advanced version, StyleGAN2~\cite{karras2020analyzing}. Despite their efforts to remove specific artifacts such as blobs, the general artifacts remain unresolved. In addition, they propose a method improving image quality, called `truncation'. This forces the style vector located near the average style vector that is calculated on training time. However, the authors state that this truncation method negatively effects the diversity of output images.

\subsection{Semantic Hierarchy Emerges in GAN}
A recent study~\cite{yang2021semantic} revealed the hierarchy mechanism of generative models. The authors of this study noted that the generation process is the same as an artist drawing a picture by interpreting it from multiple abstraction levels. In the GAN models, the early layers create the layout of the output image. The middle layers create objects and attributes. Finally, the color scheme is rendered on the late layers. This implies that the artifacts are likely to occur in the early layers.

\subsection{Detecting and removing Artifacts in Generative Models}
In recent studies, GAN Dissection~\cite{bau2018visualizing} and Automatic Correction of Internal Units in Generative Neural Networks~\cite{tousi2021automatic} have struggled to understand the inner mechanism of GANs. They suggest how the units where artifacts occur can be detedted and fixing internal errors in the GAN so that the generator can provide realistic images without artifacts. GAN Dissection removes the internal units that are expected to cause defective elements. The latter trains an auxiliary neural network classifier. For the case of an artifact image, the auxiliary classifier is used to search a region that is considered to be strange by applying Grad-CAM~\cite{2019}.

\section{Analysis for the Activation Frequency Rate}

In this section, we attempt to shed light on the causes of the diversity of generated images by observing the frequency of activations in GAN, which may lead to the generation of artifacts. We first introduce notations for GAN structure to define the frequency rate of neurons. According to the frequency rate, we verify that rarely activated neurons are highly related to unusual features and defective parts in generated images by analyzing the effects of those neurons. Furthermore, we suggest an efficient detection and ablation techniques based on our analysis.

\begin{figure}[t!]
\centering
\includegraphics[width=0.98\columnwidth]{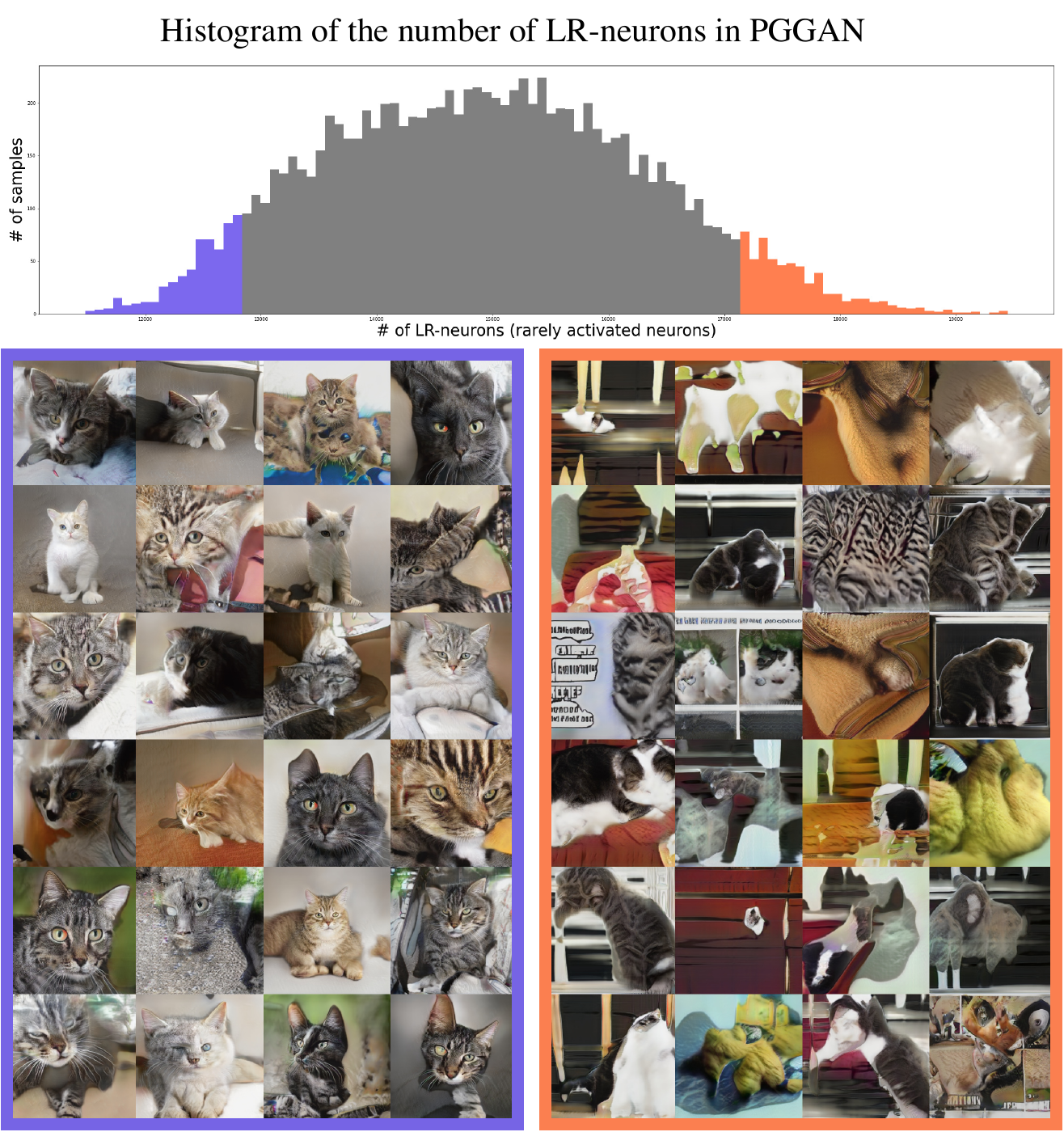} 
\caption{Generated images according to the number of rarely activated neurons in PGGAN. (Left) The samples in the blue box, of which featuremaps are composed mostly of frequently activated neurons, have general cat images. (Right) The samples in the red box have diverse components, but most of them fail to construct realistic cat images.}
\label{fig:gen_rate_hist}
\end{figure}

\subsection{The Frequency Rate of Neurons}
\paragraph{Notation.} For convenience, we introduce some notations for GAN structure and define the frequency rate to measure the frequency of activations. GAN consists of two parts, a generator $G$ and a discriminator $D$. The generator $G : \mathcal{Z} \rightarrow \mathcal{I}$ generates an output image where $\mathcal{Z}$ is the latent vector space and $\mathcal{I}$ is the space of output images. Since we analyze the process of image generation and internal neurons, we determine the generator $G$ as our main target. A function from the 0 th layer to the $i$ th layer of the generator is denoted by $G^{:i}$. The output of the $i$ th layer is denoted by $f^{i}(\mathbf{z}) = G^{(i-1):i}(f^{(i-1)}(\mathbf{z}))$, where $f^0(\mathbf{z}) = G^0(\mathbf{z})$.

\begin{definition}[\textbf{Activated Neuron}]
Given a generator $G$ and its featuremap $f^i(\mathbf{z})$ for the $i$-th layer, the activation value of the $n$-th neuron is defined by $f^i_n(\mathbf{z})$. We call it `activated' if $f^i_n(\mathbf{z}) > 0$. In particular, the set of the activated neurons in the $i$-th layer is defined by $[f^i_{\cdot}]^+$.
\end{definition}

\begin{definition}[\textbf{Neurons with Activation Rate $R$}] 
Given a featuremap $f^i$, $R^i_n$ is the probability that a neuron $f^i_n(\cdot)$ is activated. $f^i_{\cdot,R}$ is the set of neurons with an activation rate $R$ such that $R^i_n=R$.
\label{def:rate}
\end{definition}

The activation rate $R^i_n$ denotes how frequently the neuron $n$ is indeed activated:
\begin{equation}
    R^i_n = \mathbb{E}_{\mathbf{z}\sim \mathcal{Z}} [I(f^i_n(\mathbf{z}) > 0)]
\end{equation}

where $I$ is the indicator function. Since the true probability for each neuron cannot be computed exactly, 30K latent codes are sampled to estimate the relative frequency $R^i_n$. We analyze the effect of neurons with low frequency rates and that of neurons with high frequency rates. Then, with the constant rate $R$, we denote the set of neurons which have $R^i_n>R$ by $\overline{f}^i_{\cdot,R}$ and denote the set of neurons that have $R^i_n \le R$ by $\underline{f}^i_{\cdot,R}$. When there is no confusion in the context, we simply call $\overline{f}^i_{\cdot,R}$ `HR-neurons' (high-rate neurons) and call $\underline{f}^i_{\cdot,R}$ `LR-neurons' (low-rate neurons).

\begin{figure}[b!]
\centering
\includegraphics[width=0.95\columnwidth]{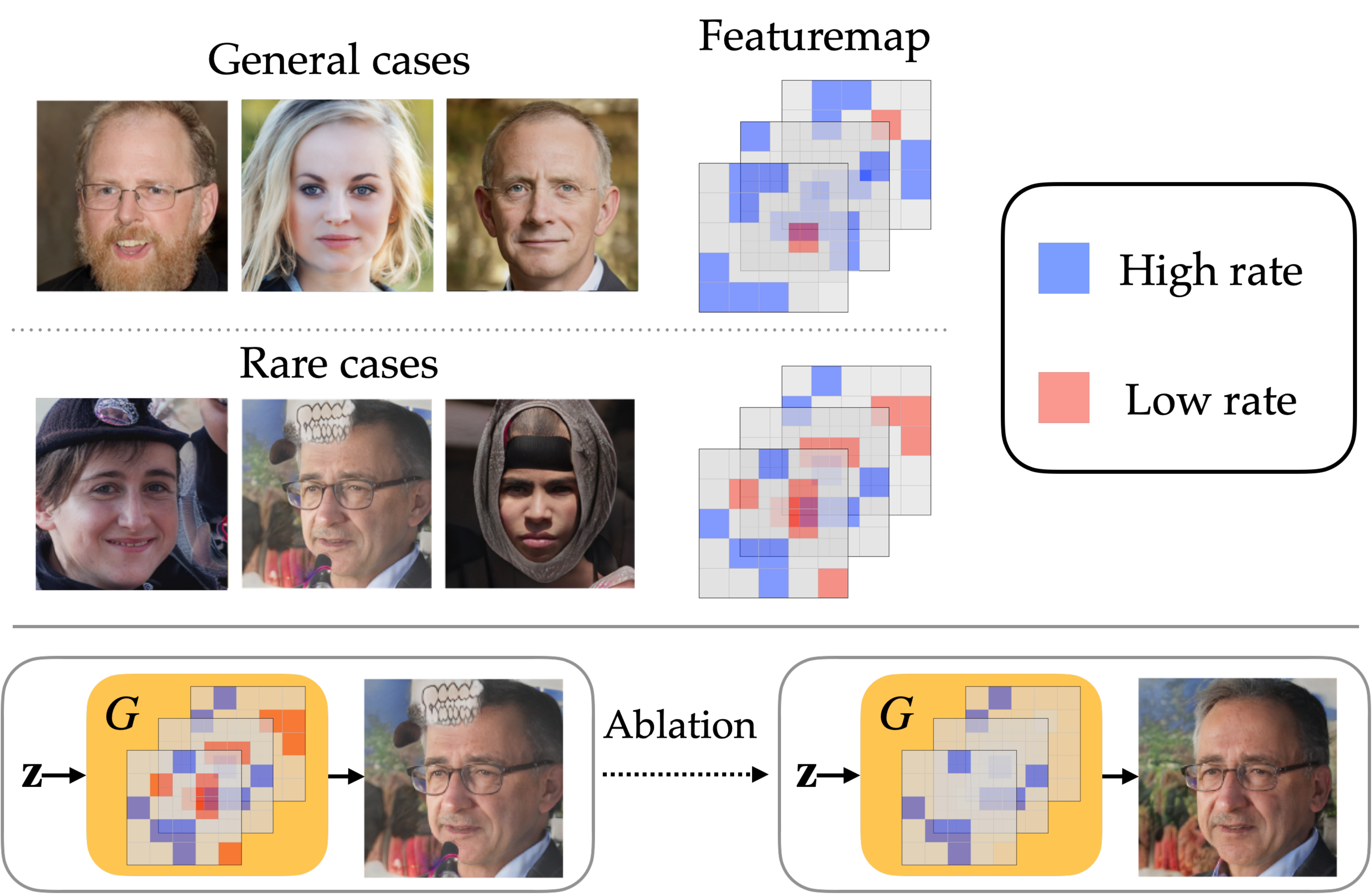} 
\caption{ (Top) Comparison of feature maps between general samples and rare samples and (Bottom) our proposed ablation process. The red color denotes rarely activated neurons and the blue color denotes frequently activated neurons.}
\label{fig:seq_abl}
\end{figure}

\paragraph{Hypothesis.} In featuremaps of generated images, some neurons are rarely activated whereas other neurons are frequently activated. In other words, neurons have different activation frequency, which are precisely defined as the `frequency rate' in Definition~\ref{def:rate}. We observe that featuremaps of normal images among generated results are composed mostly of frequently activated neurons. For instance, the rare cases shown in Figure~\ref{fig:seq_abl} have many neurons with low frequency rates in their featuremaps unlike general cases. Considering this, we hypothesize that rarely activated neurons are used to create diverse components which may include failures, but frequently activated neurons are used to construct general components that are necessary for realistic images.  

Figure~\ref{fig:gen_rate_hist} provides empirical evidences of our hypothesis. The histogram shows the number of generated images according to the number of activated neurons that are rarely activated during the generating procedure. The images in the red box, of which featuremaps have many rarely activated neurons, have diverse components, such as images with phrases. Most of them, however, fail to construct realistic images, while the images in the blue box show general cat images. It is reasonable to suspect that rarely activated neurons are highly related to artifact images generated from GANs.

\begin{algorithm}[ht!]
\caption{Sequential Ablation}
\label{seq}
\textbf{Input}: Generator $G$, A latent code $\mathbf{z} \in \mathcal{Z}$\\
\textbf{Parameter}: All Layer index list $\mathbf{L}$, Layer index list $\mathbf{M}$ for ablation, Activation rate $R$\\
\textbf{Output}: Corrected Image $I_{corr}$
\begin{algorithmic}[1] 
\STATE Let $I \leftarrow \mathbf{z}$.
\FOR{ i in  $\mathbf{L}$}
\STATE $f^i \leftarrow G^{(i-1):i}(I)$
\IF {i $\mathbf{in}$ $\mathbf{M}$}
\STATE $\big[\underline{f}^i_{\cdot,R}(I)\big]^+ \leftarrow 0$
\ENDIF
\STATE $I \leftarrow f^i$
\ENDFOR
\STATE $I_{corr} \leftarrow I$
\STATE \textbf{return} $I_{corr}$
\end{algorithmic}
\label{alg:seqab}
\end{algorithm}

\subsection{Relationship between Frequency Rate and Generations}

To observe the effects of neurons according to their frequency rates, we first introduce an ablation method, namely `Sequential Ablation’, in Algorithm~\ref{alg:seqab}. Given a latent code $\mathbf{z}\in\mathcal{Z}$, it iteratively computes its featuremap while turning off activated LR-neurons for each layer. In other words, it makes values of activated LR-neurons as zeros. The target layers to remove LR-neurons are limited to indices in \textbf{M}, since neurons on different layers may have different effects on generated images. Previous research shows that each layer in GANs plays a different role in the generating process in a hierarchical manner~\cite{yang2021semantic}. The authors note that the early and middle layers determine the spatial layout and the objects, and the last layers render the attributes and color scheme. Therefore, we set \textbf{M} to include the early or middle layers of the generator, since our work focuses on analyzing the root of unrealistic objects or defective parts.


\subsubsection{Analysis for Layer}
\begin{figure}[t!]
\centering
\includegraphics[width=0.94\columnwidth]{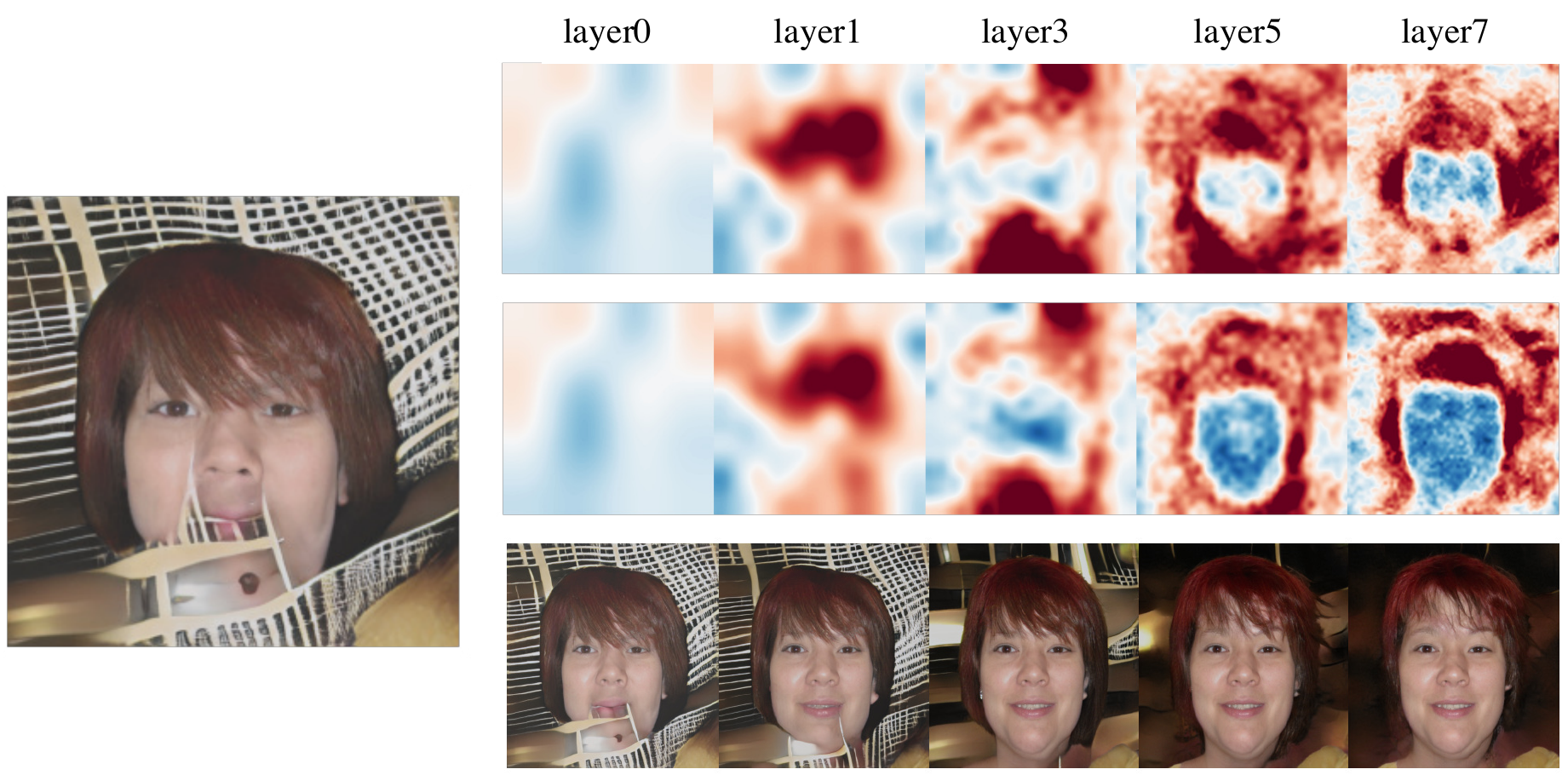} 
\caption{Heatmaps of neurons with low frequency rate and their effects on generations. The left most column is generated artifact image. The first row shows the masks of activations corresponding to $[\underline{f}^i_{\cdot,0.3}(\mathbf{z})]^+$. The second row shows the masks of activations that are applied sequential ablation upto a given layer. The third row describes the generated images after sequential ablation.}
\label{fig:layers_mask}
\end{figure}

For a randomly selected artifact image generated from StyleGAN2, the first row in Figure~\ref{fig:layers_mask} shows how LR-neurons are distributed in its featuremap for each layer. The generated image has a strange object on its chin and a grid-like background. The red part of the heatmaps indicates a high proportion of LR-neurons. The heatmaps on layers 5 and 7 have a large number of LR-neurons at the chin. The second row shows heatmaps and the third row shows generated images, when applying sequential ablation until the previous layers. In other words, the heatmap of layer 5 in the second row denotes the heatmap when applying sequential ablation at layers 0, 1, and 3. Through sequential ablation, LR-neurons at the defective parts of the chin are removed and the generated image is also successfully repaired. Therefore, we can identify that LR-neurons in the early layers are highly related to causing defective parts. However, when we apply sequential ablation for the higher layers, most of the background part disappears and the hairstyle become different. This implies that LR-neurons in layers 5 and 7 contributes to constructing detail features or the background. From this result, we conclude that applying sequential ablation until layer 3 is an appropriate choice to repair generated images without hindering the diversity of generations. In Section~\ref{sec:quant}, the artifact correction results in Table~\ref{tab:FID} support our choice quantitatively.

\subsubsection{Analysis for Rate}
\label{sec:anal_rate}

\begin{figure*}[ht!]
\centering
\includegraphics[width= 0.95\textwidth]{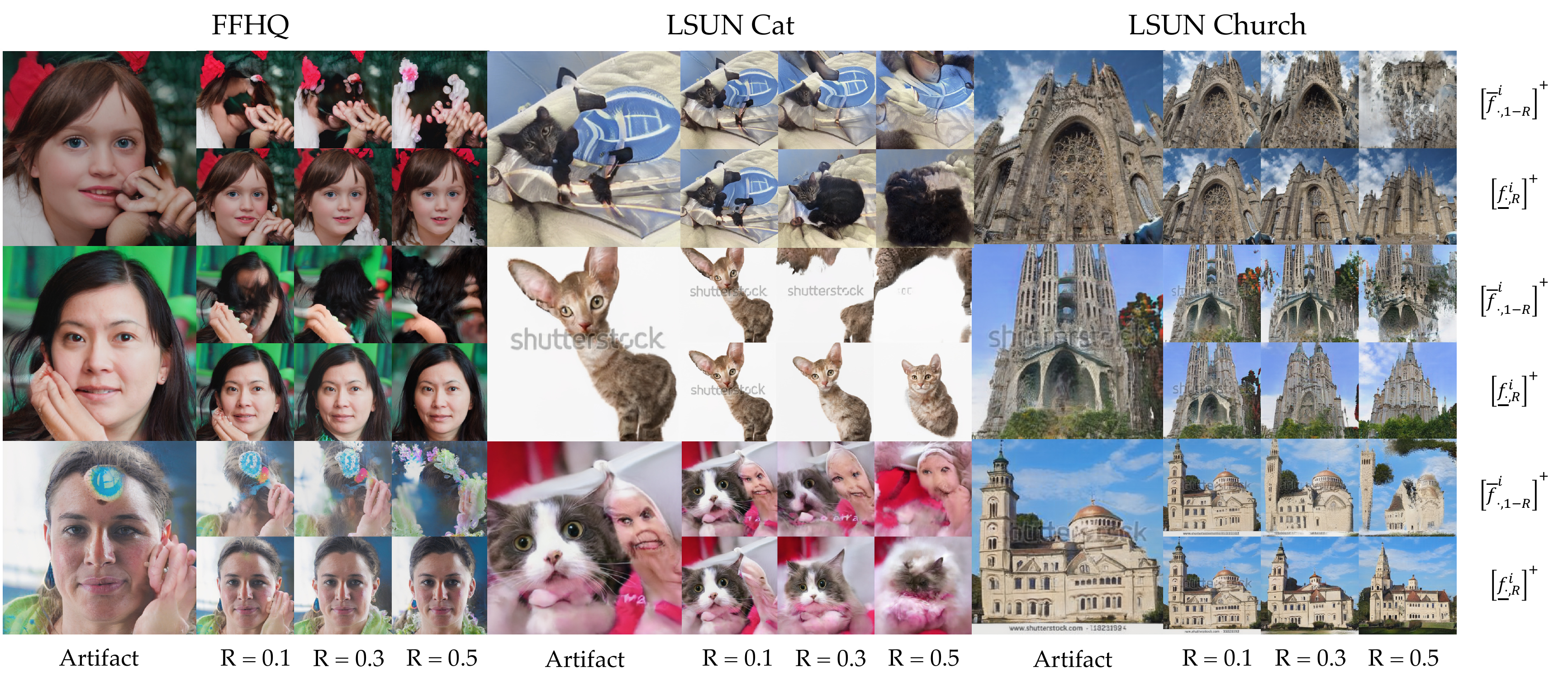} 
\caption{Generated images from StyleGAN2 with ablating HR-neurons or LR-neurons. For each sample, the first row shows the results of ablating HR-neurons, and the second row shows the results of ablating LR-neurons.}
\label{fig:diff-ab-samples}
\end{figure*}

In this section, we provide qualitative analysis to verify our hypothesis and reveal the roles of neurons according to the frequency rate. Figure~\ref{fig:diff-ab-samples} shows the generation results of StyleGAN2 with sequential ablation applied in layers 0, 1, and 3 with different rates on FFHQ, LSUN-Cat,and LSUN-Church. In this analysis, we remove not only LR-neurons but also HR-neurons to identify the effects of HR-neurons even though sequential ablation originally denotes ablating LR-neurons. For each generated image, the first row shows results from ablating HR-neurons with a given $R$ and the second row shows results from ablating LR-neurons.

When ablating HR-neurons, the basic structures of the objects, such as the shapes of faces, totally collapse. Uncommon characteristics, such as the blue artifact in the first cat image and the human-like artifact in the third cat image, still remain after ablating HR-neurons. Even though the remaining neurons, which are rarely activated, may be necessary parts to generate images with diverse properties, they easily hinder visual fidelity.

Artifacts in the generated images are removed when LR-neurons are ablated. Distorted hands in the generated human images successfully disappear with $R=0.3$ or $0.5$. The watermarks in the second cat image and the second church image are eliminated, and their shapes become more natural. However, ablating $[\underline{f}^i_{\cdot, 0.5}]^+$ may remove too much information so that generated images have only fundamental components and lose their properties. In this case, $R=0.3$ is a valid choice for sequential ablation to remove defective parts while maintaining the details.

\subsection{Ablating Neurons to Repair Artifacts}
\label{sec:repair}
From our previous analysis, we observe that LR-neurons in the early layers are highly related to generate defective parts in the results. Considering this, we propose an efficient ablation method to avoid generating artifact images. It follows the `Sequential Ablation' method, which is described in Algorithm~\ref{alg:seqab}, and the early layers are set as a list of target layers $M$ with $R=0.3$. Experimental results in Section~\ref{sec:quant} will show that this setting is valid.

Figure~\ref{fig:seq_abl} describes the overall correction process. Otherwise, we denote applying ablation is only denoted for a single layer by `Single Ablation’. In Table~\ref{tab:FID}, we use single ablation to compare the performance. Note that our ablation method does not require complex computations or manual analysis in the repair process. We can easily repair the image by setting activated LR-neurons as zero.

{\renewcommand{\arraystretch}{1.3}
\begin{table}[t!]
\centering
\resizebox{\columnwidth}{!}{
{\Large
\begin{tabular}{lrrrrrrr}
\hline
\textbf{Layers}                                  & None                    & 0             & 1             & 3             & 0,1           & 0,1,3         & 0,1,3,5       \\ \hline
{$\big[\underline{f}^i_{\cdot,0.7}\big]^+$} &  & 240.8         & 241.2         & 259.0         & 233.3         & 277.7         & 342.5         \\ 
{$\big[\underline{f}^i_{\cdot,0.5}\big]^+$} &                       &\textbf{36.3}  & \textbf{31.6} & 33.4          & \textbf{30.9} & 38.5          & 56.7          \\
{$\big[\underline{f}^i_{\cdot,0.3}\big]^+$} & {48.9}     & 44.9          & 37.0          & \textbf{31.9} & 35.6          & \textbf{29.6} & 39.2          \\ 
{$\big[\underline{f}^i_{\cdot,0.1}\big]^+$} &                       & 48.7          & 46.9          & 41.5          & 46.8          & 40.3          & \textbf{38.3} \\ 
{$\big[\overline{f}^i_{\cdot,0.7}\big]^+$}  &                       & 74.1          & 124.1         & 118.6         & 147.2         & 155.1         & 163.0         \\  
\text{Random}                                             &                       & 43.5          & 44.5          & 55.4          & 52.4          & 178.0         & 226.7         \\ \hline
\end{tabular}
}
}
\caption{FID scores for repaired images with various target layer indices and rate $R$ in StlyeGAN2 with the FFHQ dataset.}
\label{tab:FID}
\end{table}}

\section{Experiment}
\label{sec:exp}

In this section, we carry out the following qualitative and quantitative experiments to evaluate our approach. The main results show the relationship between LR-neurons and low visual fidelity (or artifacts). Deep generative models were reproduced in PyTorch from genforce github~\cite{genforce2020} which provides the official pre-trained weights; PGGANs trained on CelebA-HQ, LSUN-Cat, and LSUN-Church and StyleGAN2s trained on FFHQ, LSUN-Cat, and LSUN-Church.

\begin{figure*}[t!]
\centering
\includegraphics[width=0.95\textwidth]{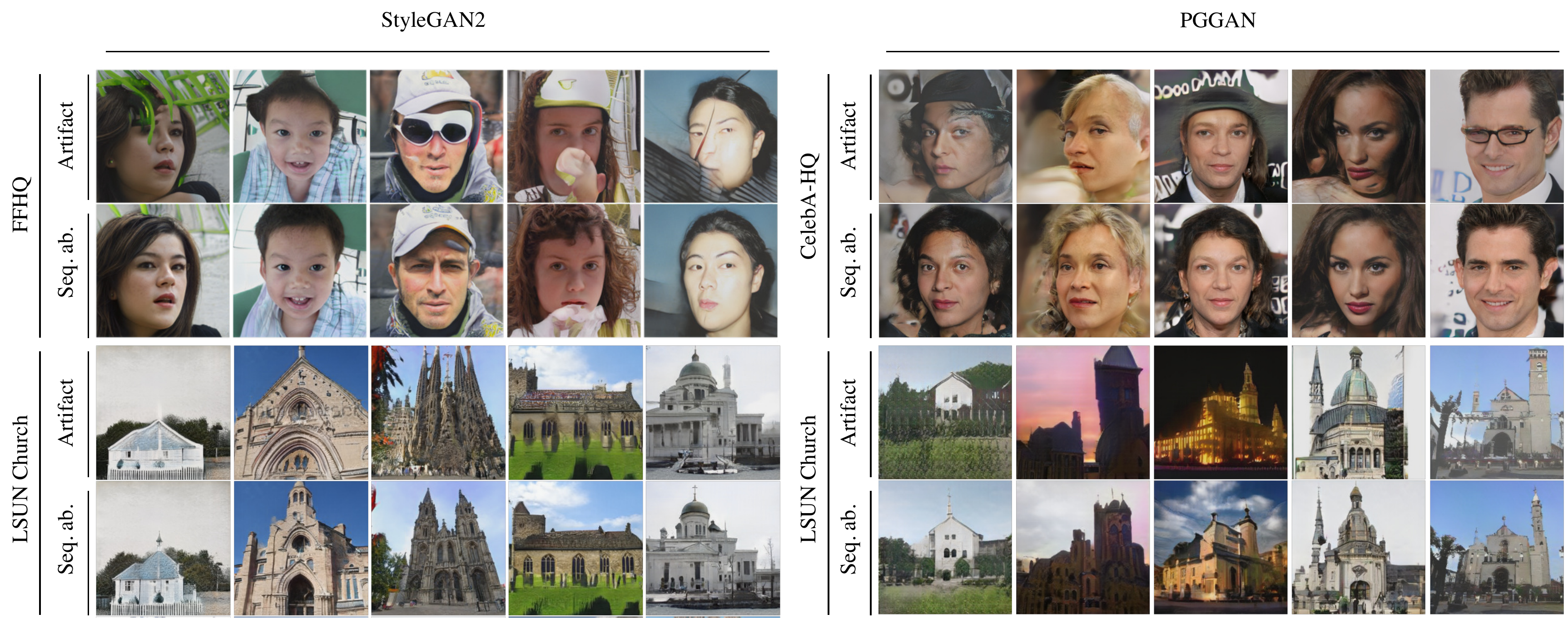} 
\caption{Sequential Ablation results for correcting artifacts in StyleGAN2s and PGGANs}
\label{fig:corr-samples}
\end{figure*}

\begin{table*}[h!]
\centering
\resizebox{0.9\textwidth}{!}{
\begin{tabular}{c|c|rrr|rrr}
\hline
\multirow{2}{*}{}          & \multirow{2}{*}{Datasets} & \multicolumn{3}{c|}{Bottom 2K}                                                                                                                                                          & \multicolumn{3}{c}{Top 2K}                                                                                                                                                   \\ \cline{3-8} 
                          &                           & \multicolumn{1}{c|}{\begin{tabular}[c]{@{}c@{}}Precision\\ (fidelity)\end{tabular}} & \multicolumn{1}{c|}{\begin{tabular}[c]{@{}c@{}}Recall\\ (diversity)\end{tabular}} & Realism Score           & \multicolumn{1}{c|}{\begin{tabular}[c]{@{}c@{}}Precision\\ (fidelity)\end{tabular}} & \multicolumn{1}{c|}{\begin{tabular}[c]{@{}c@{}}Recall\\ (diversity)\end{tabular}} & Realism Score  \\ \hline
\multirow{3}{*}{PGGAN}     & CelebA-HQ                 & \textbf{0.9245}                                                                     & 0.1056                                                                            & \textbf{1.0911$\pm$0.0712} & 0.4565                                                                              & \textbf{0.3194}                                                                   & 0.9942$\pm$0.0663 \\
                          & LSUN-Cat                  & \textbf{0.8450}                                                                     & 0.0893                                                                            & \textbf{1.0680$\pm$0.0686} & 0.4745                                                                              & \textbf{0.1089}                                                                   & 1.0011$\pm$0.0571 \\
                          & LSUN-Church               & \textbf{0.8985}                                                                     & 0.1907                                                                            & \textbf{1.0754$\pm$0.0627} & 0.4165                                                                              & \textbf{0.3628}                                                                   & 0.9902$\pm$0.0697 \\ \hline
\multirow{3}{*}{StyleGAN2} & FFHQ                      & \textbf{0.9085}                                                                     & 0.2153                                                                            & \textbf{1.1112$\pm$0.0913} & 0.762                                                                               & \textbf{0.4763}                                                                   & 1.0370$\pm$0.0725 \\
                          & LSUN-Cat                  & \textbf{0.9195}                                                                     & 0.1404                                                                            & \textbf{1.0875$\pm$0.0704} & 0.6385                                                                              & \textbf{0.4025}                                                                   & 1.0206$\pm$0.0579 \\
                          & LSUN-Church               & \textbf{0.8315}                                                                     & 0.2226                                                                            & \textbf{1.0631$\pm$0.0640}  & 0.4935                                                                              & \textbf{0.3646}                                                                   & 1.0016$\pm$0.0729 \\ \hline
\end{tabular}
}
\caption{Precision, Recall, and Realism Score of Top/Bottom images with a lot/few of LR-neurons.}
\label{tab:precision, recall, RS}
\end{table*}

\subsection{Qualitative Results}

50K latent codes were randomly sampled to generate images without the truncation method. The number of the total activated LR-neurons ($R=0.3$) in the early layers was counted for each generated image, i.e., the sum of $\big | \big[\underline{f}^i_{\cdot,0.3} \big]^+ \big |$. The target early layers were set to $i=1,3,5$ for PGGAN and $i=0,1,3$ for StyleGAN2. The dimensionality of each layer's output corresponds to $\mathbb{R}^{512 \times 4 \times 4},\mathbb{R}^{512 \times 8 \times 8}, \text{ and } \mathbb{R}^{512 \times 16 \times 16}$, respectively.

\subsubsection{Low Fidelity Detection in Image Generations}
To detect images with low visual fidelity, the generated images were sorted according to the number of activated LR-neurons in decreasing order. the top 2K images and Bottom 2K images are then selected. As an example, Figure~\ref{fig:gen_rate_hist} shows the histogram of the number of activated LR-neurons. The blue region indicates that they have few activated LR-neurons and the red region indicates that the generated images have many activated LR-neurons. Consistent with our hypothesis, the generated images in the red box have more defective or distorted parts than the images in the blue box. The images in the blue box have a high visual fidelity, but lack diversity, whereas the images in the red box seem to have more uncommon properties. We provide abundant results of detection tasks in Appendix B-G.

\subsubsection{Sequential Ablation}
Figure~\ref{fig:diff-ab-samples} shows sequential ablation results of StyleGAN2 in FFHQ, LSUN-Cat, and LSUN-Church. The ablation method is conducted on layer 0, 1, and 3 with various rate $R$ as described in Section~\ref{sec:anal_rate}. This supports that the LR-neurons induce the artifacts since removing them helps to generate realistic images. Furthermore, in Section~\ref{sec:repair}, sequential ablation is conducted to correct artifacts generated from StyleGAN2 and PGGAN. As shown in Figure~\ref{fig:corr-samples}, we identify that our approach can repair the artifacts in various cases. In the case of StyleGAN2, a long crack on the face of the fifth human image was successfully eliminated. The buildings are not completely constructed and mixed with their backgrounds in the artifact church images generated from PGGAN. After sequential ablation, the buildings are repaired normally. 

On the unaligned or noisy datasets such as LSUN, however, the HR-neurons have relatively less responsibility for creating a single representative object in the dataset. It is because the datasets that have sparse features can confuse the generative model to consider the objects to be mainly created as part of diversity. Therefore, our approach may work better on the well-aligned datasets. Additional results are provided in Appendix H (including the results of StyleGAN3~\cite{karras2021alias} with FFHQ).

\subsection{Quantitative Results}
\label{sec:quant}

\subsubsection{Correction Study}
Here, we evaluate our ablation method with various combinations of target layers and rate $R$ in the FFHQ dataset with FID score~\cite{heusel2017gans}. Table~\ref{tab:FID} shows FID scores computed with 10K real images and 1.3K generated artifacts images. Examples of hand labeled generated images are in Appendix A. `None' denotes a case when we do not apply our ablation metho\textbf{}d and `Random' denotes when we ablate 30\% of randomly selected activated neurons. Note that we remove activated HR-neurons to identify the effects of HR-neurons on generations in the case of the 5th row. The ablation result when turning off $[\underline{f}^i_{\cdot, 0.3}]^+$ with $i = 0,1,3$ has the best FID score. Compared to the case $[\underline{f}^i_{\cdot, 0.3}]^+$, the case $[\overline{f}^i_{\cdot, 0.7}]^+$ has very high FID scores because HR-neurons play a significant role in generating essential parts of images. The most optimal setting is used ($[\underline{f}^i_{\cdot, R}]^+$ with $R = 0.3, i = 0,1,3$) to conduct our qualitative experiments.

\subsubsection{Diversity and Fidelity}
We also quantify our detection method with the precision (fidelity), recall (diversity), and Realism Score~\cite{NEURIPS2019_0234c510} with 10K real images, the Top 2K and Bottom 2K samples for each dataset, and the number of neighborhood $k=3$. Table~\ref{tab:precision, recall, RS} shows that the Bottom 2K samples that contain few LR-neurons have a high score in fidelity and realism score, but a low score in recall. On the other hand, the fidelity and realism score for the Top 2K samples that contain a large number of LR-neurons are lower than those of the Bottom 2K samples, while diversity is higher in all cases. These results are linked to the evidence of our hypothesis that encouraging diversity leads to artifacts. In addition, it may imply a trade-off between fidelity and diversity in the generation procedure.

\section{Conclusion}
In this paper, we suggest a hypothesis that rarely activated neurons are highly related to generating artifacts, which might be induced as failures to create diversity. Our ablation study not only supports that low-rate neurons in the early layers induce defective parts but also shows that frequently activated neurons may affect essential parts. As a result, our approach can detect artifact images and correct defective parts by sequential ablation without any supervision.

\section*{Acknowledgements}
This work was supported in part by the Institute of Information and Communications Technology Planning and Evaluation (IITP) grant funded by the Korea Government (MSIT) under Grant 2019-0-00075, Grant 2022-0-00184, and Grant 2022-0-00984 and partly supported by KAIST-NAVER Hypercreative AI Center.

\bibliographystyle{named}
\bibliography{ijcai22}
\newpage

\onecolumn

\input{appendix/appendix}

\end{document}

%% file: appendix/appendix.tex
\pdfinfo{
}





\section*{\centering \textbf{APPENDIX}}

\subsection*{\centering A. Hand labeled Normal/Artifact images in StyleGAN2 with FFHQ}

\begin{figure*}[h!]
\centering
\includegraphics[width=0.92\textwidth]{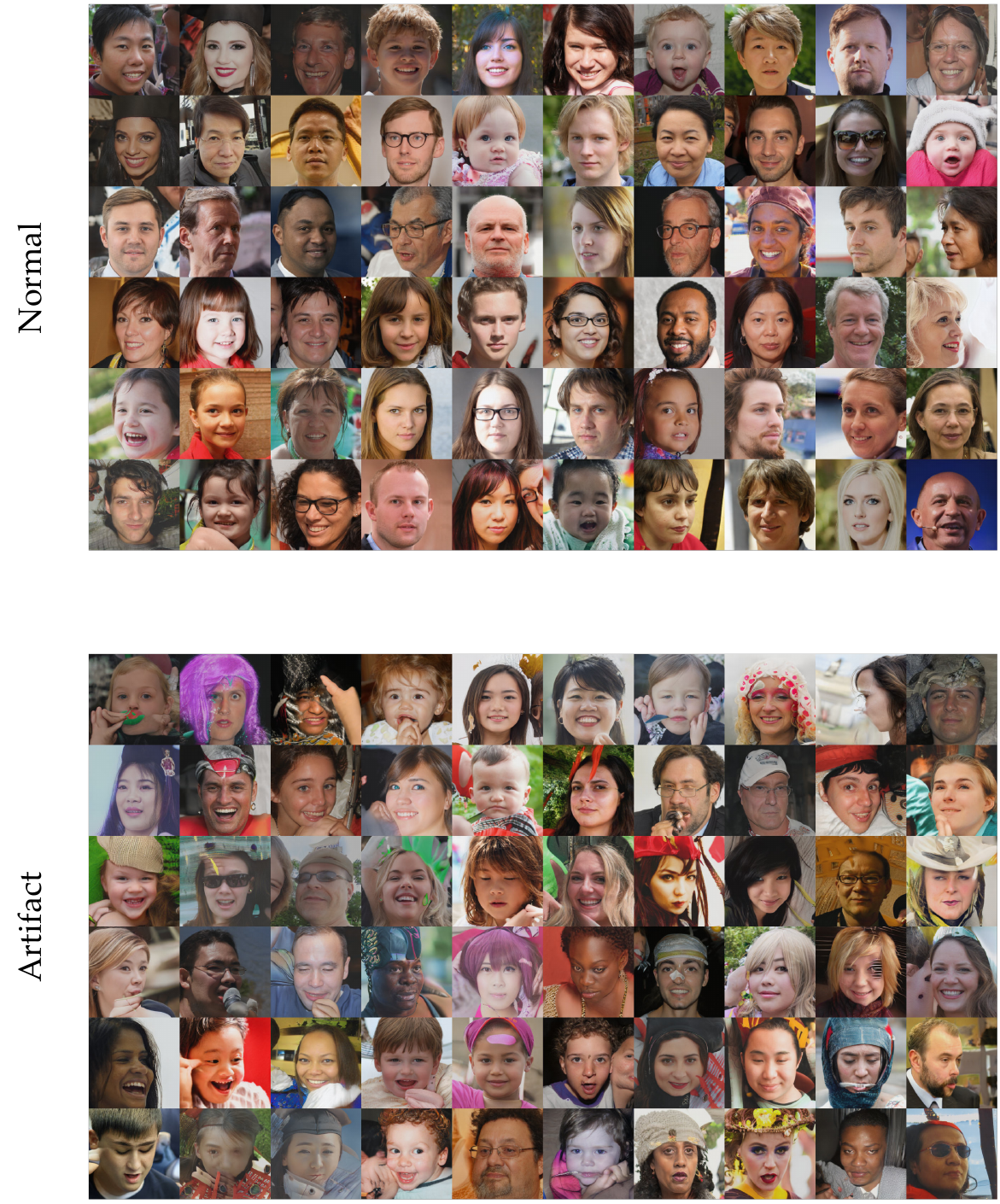} 
\caption{We labeled total 10K images and show the examples of uncurated labeled images. We classified it 'Artifact' if it has distorted/defective part on face or weird objects which are look like hat, hand, and mic. Otherwise, We classified it 'Normal'. (Normal : 8.7K, Artifact : 1.3K)}
\label{fig:diff-ab-samples}
\end{figure*}

\clearpage

\subsection*{\centering B. Artifact detection in PGGAN with CelebA-HQ}

\begin{figure*}[h!]
\centering
\includegraphics[width=0.92\textwidth]{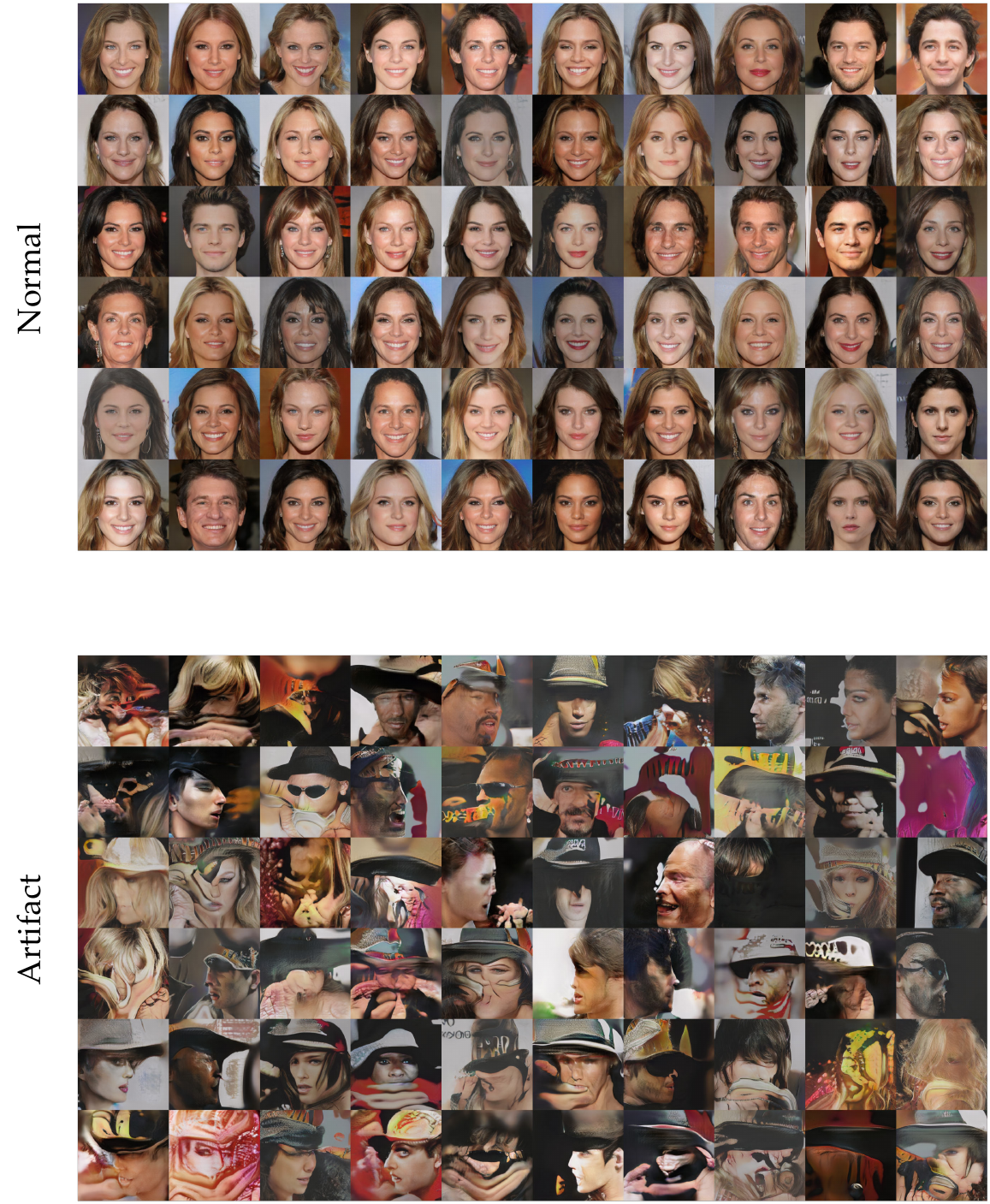} 
\caption{For 50K random samples, the images in normal are the bottom 60 images and the images in artifact corresponds to the top 60, in decreasing order with the number of LR-neurons.}
\label{fig:diff-ab-samples}
\end{figure*}

\clearpage

\subsection*{\centering C. Artifact detection in PGGAN with LSUN-Cat}

\begin{figure*}[h!]
\centering
\includegraphics[width=0.92\textwidth]{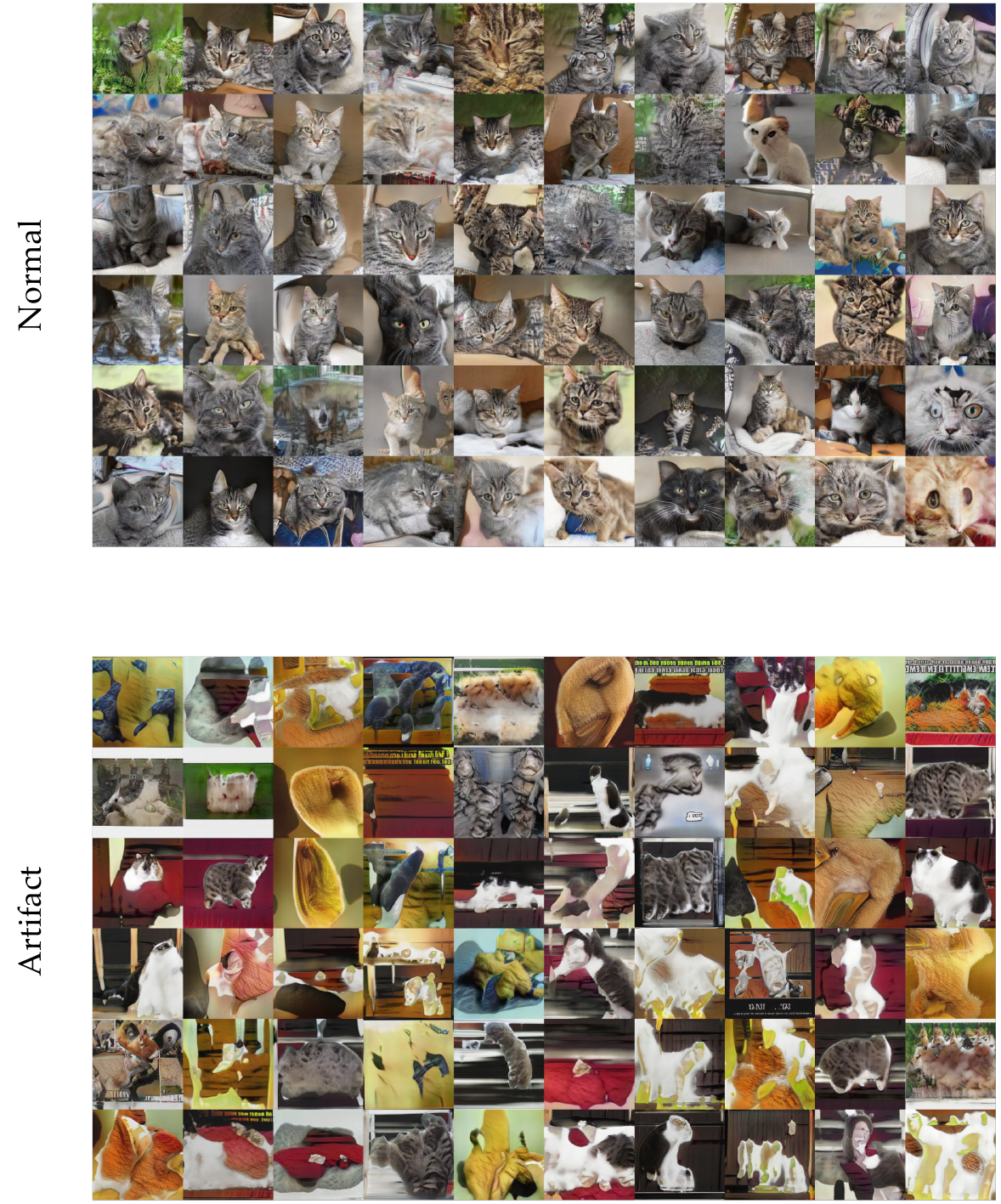} 
\caption{For 50K random samples, the images in normal are the bottom 60 images and the images in artifact corresponds to the top 60, in decreasing order with the number of LR-neurons.}
\label{fig:diff-ab-samples}
\end{figure*}

\clearpage

\subsection*{\centering D. Artifact detection in PGGAN with LSUN-Church}

\begin{figure*}[h!]
\centering
\includegraphics[width=0.92\textwidth]{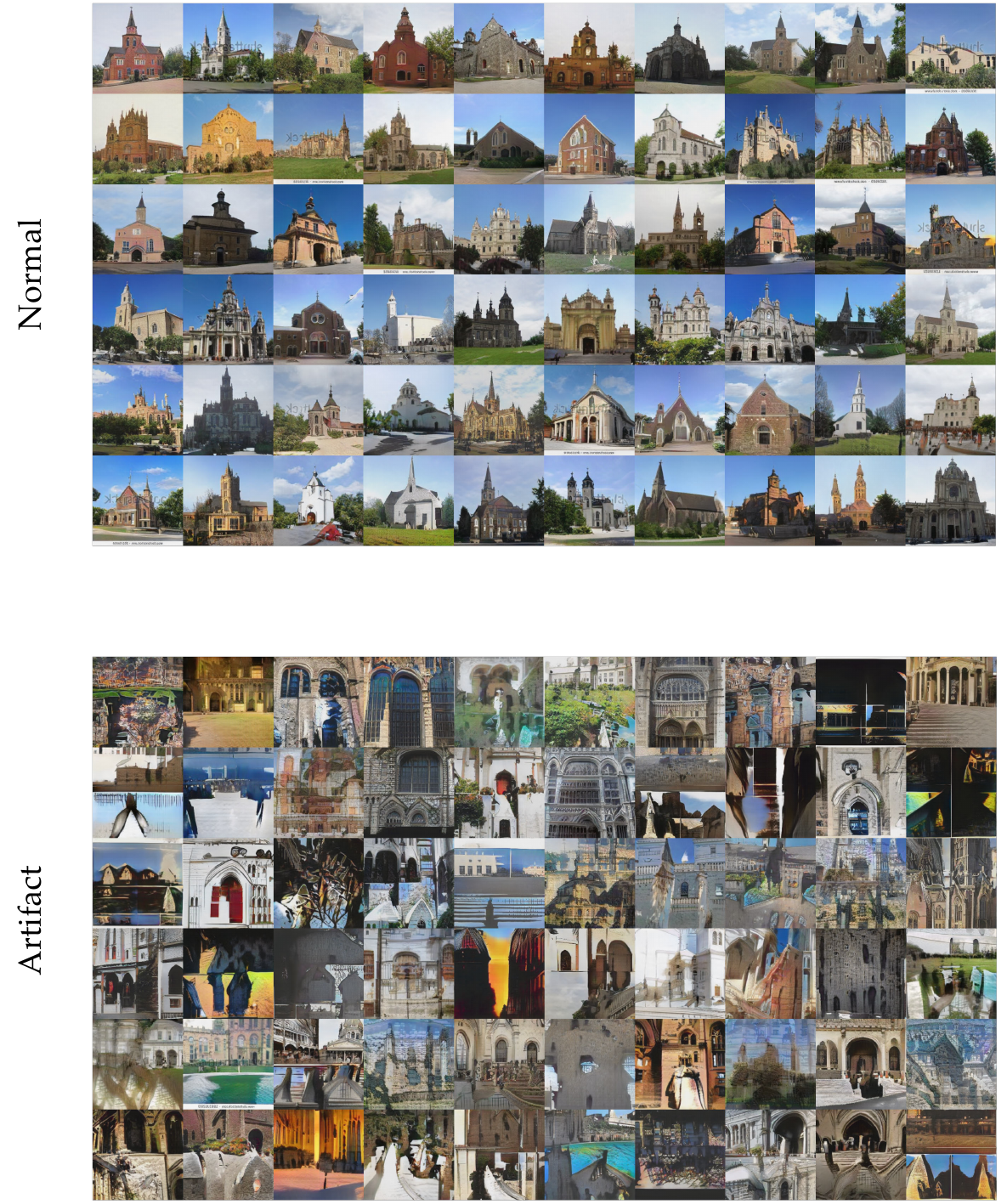} 
\caption{For 50K random samples, the images in normal are the bottom 60 images and the images in artifact corresponds to the top 60, in decreasing order with the number of LR-neurons.}
\label{fig:diff-ab-samples}
\end{figure*}

\clearpage

\subsection*{\centering E. Artifact detection in StyleGAN2 with FFHQ}

\begin{figure*}[h!]
\centering
\includegraphics[width=0.92\textwidth]{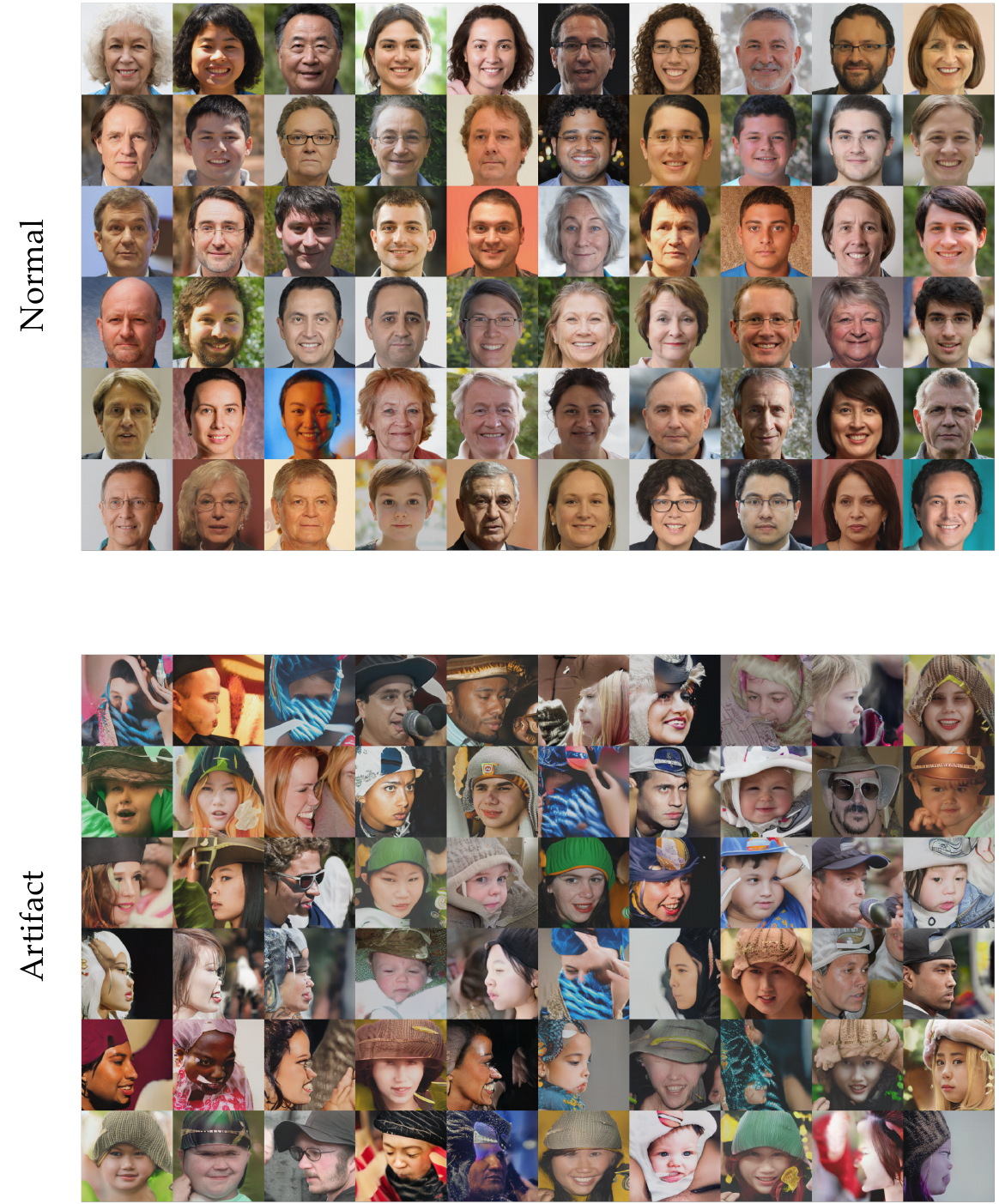} 
\caption{For 50K random samples, the images in normal are the bottom 60 images and the images in artifact corresponds to the top 60, in decreasing order with the number of LR-neurons.}
\label{fig:diff-ab-samples}
\end{figure*}

\clearpage

\subsection*{\centering F. Artifact detection in StyleGAN2 with LSUN-Cat}

\begin{figure*}[h!]
\centering
\includegraphics[width=0.92\textwidth]{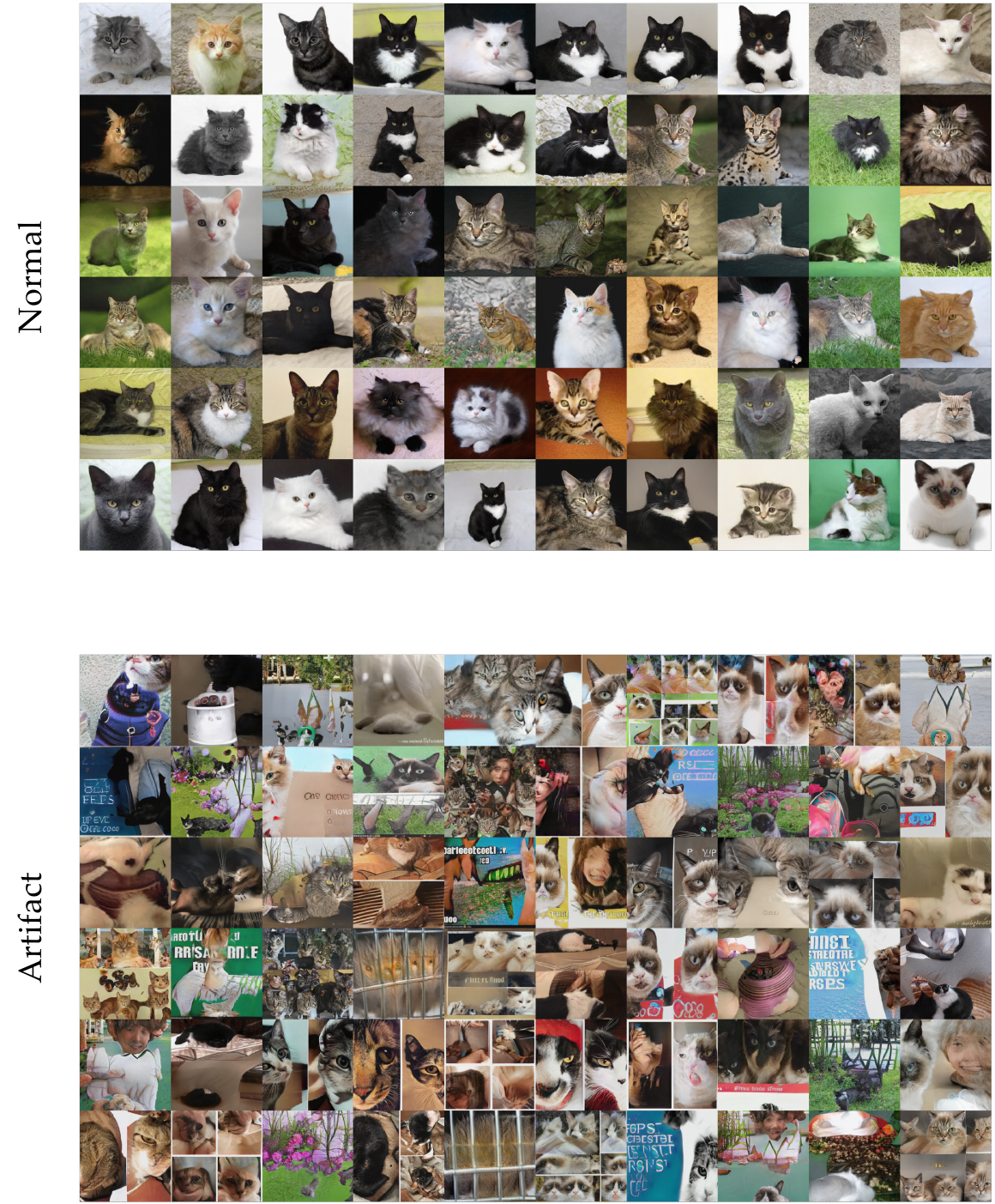} 
\caption{For 50K random samples, the images in normal are the bottom 60 images and the images in artifact corresponds to the top 60, in decreasing order with the number of LR-neurons.}
\label{fig:diff-ab-samples}
\end{figure*}

\clearpage

\subsection*{\centering G. Artifact detection in StyleGAN2 with LSUN-Church}

\begin{figure*}[h!]
\centering
\includegraphics[width=0.92\textwidth]{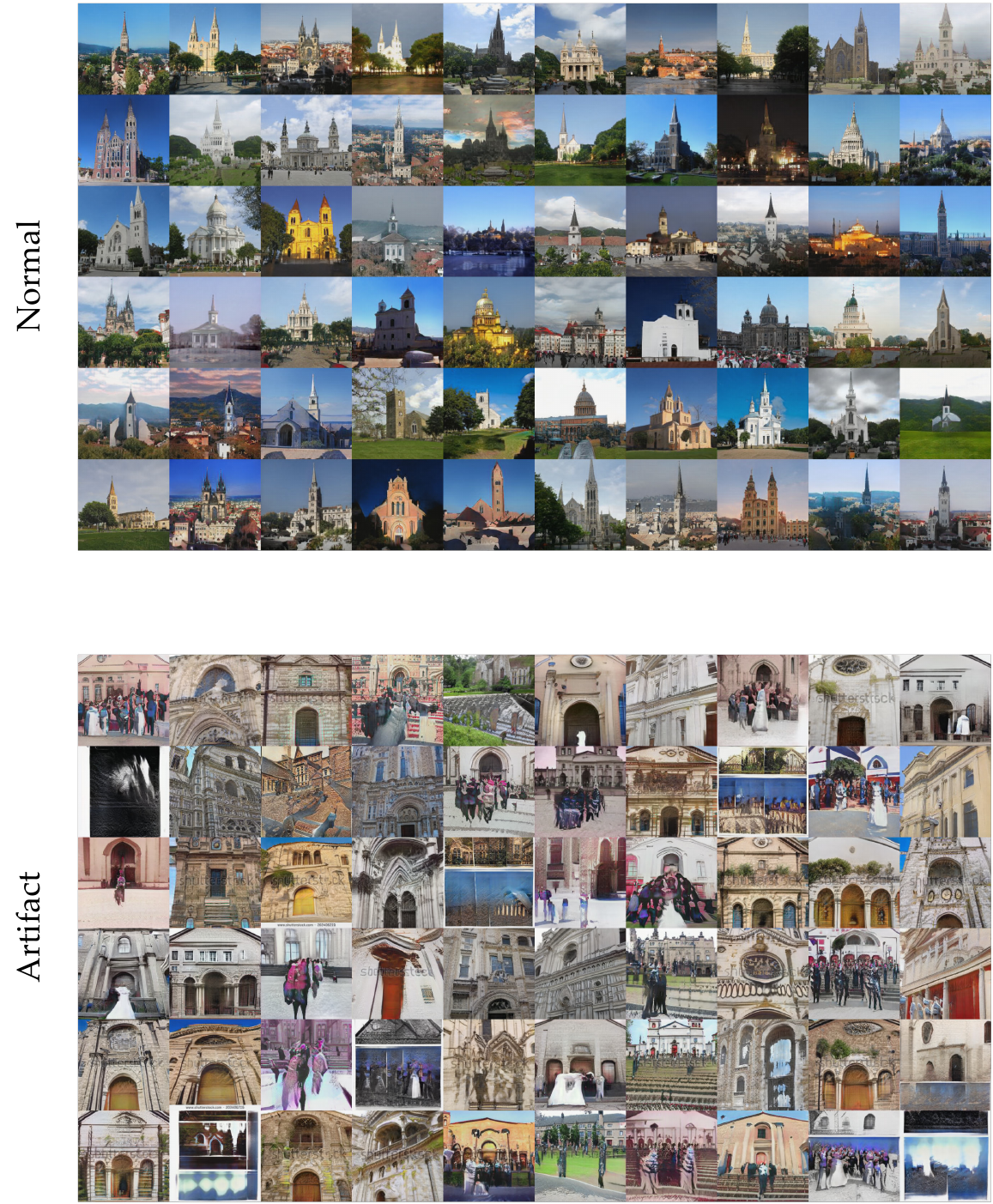} 
\caption{For 50K random samples, the images in normal are the bottom 60 images and the images in artifact corresponds to the top 60, in decreasing order with the number of LR-neurons.}
\label{fig:diff-ab-samples}
\end{figure*}

\clearpage

\subsection*{\centering H. Sequential Ablation to remove artifact}

\begin{figure*}[h!]
\centering
\includegraphics[width=0.77\textwidth]{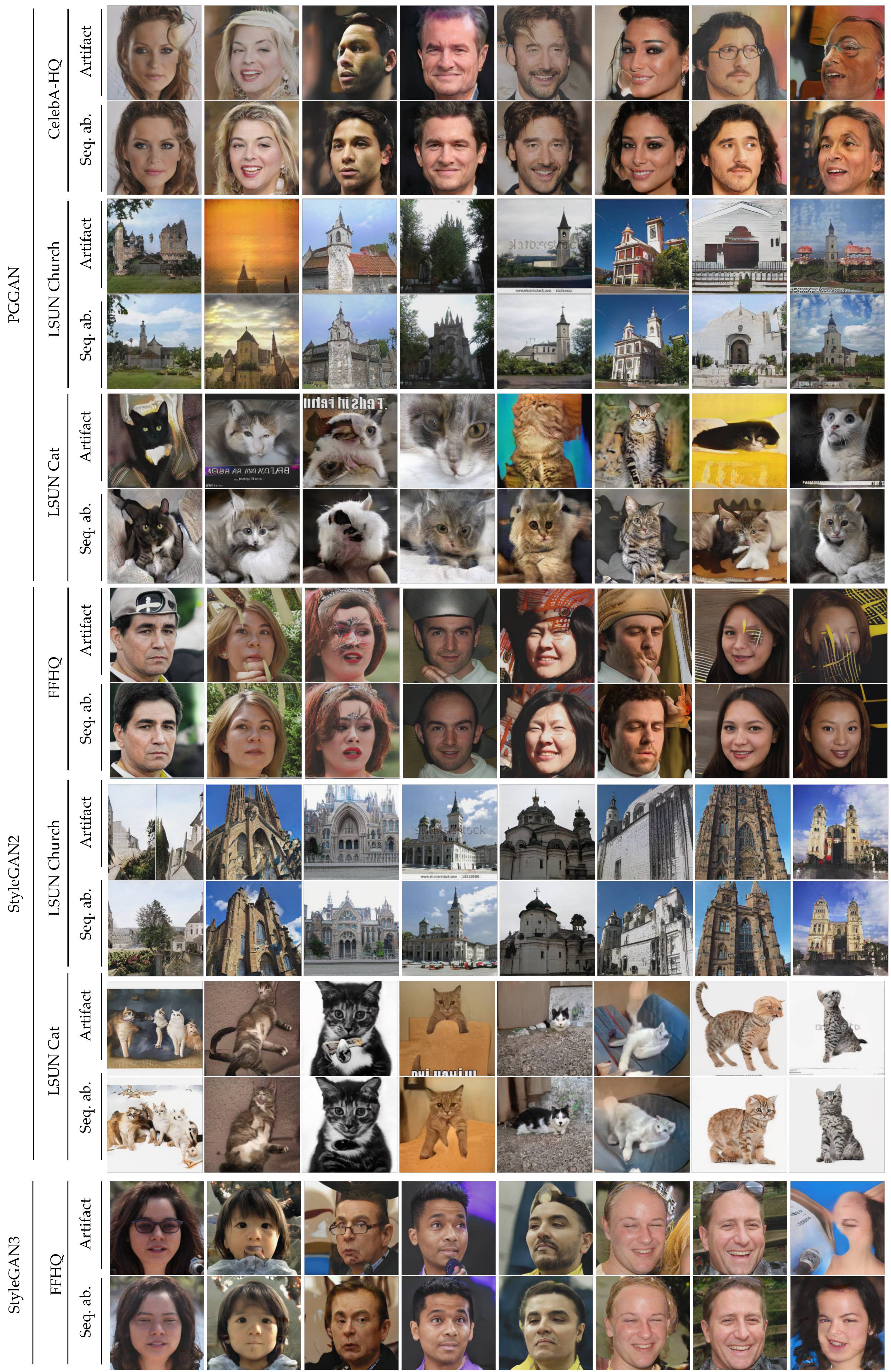} 
\caption{Sequential ablation with $R=0.3$ in various datasets and models}
\label{fig:diff-ab-samples}
\end{figure*}
